\documentclass[sigconf]{acmart}
\usepackage{algorithm}
\usepackage{algorithmic}  
\usepackage{multirow}
\usepackage{xcolor} 
\definecolor{customcolor}{RGB}{235, 235, 235} 
\definecolor{green}{RGB}{ 0,136,51}
\definecolor{blue}{RGB}{0,102,204}
\definecolor{red}{RGB}{ 202,0,0}
\definecolor{mysuppcolor}{RGB}{197,116,23}
\definecolor{suppcolor}{RGB}{0,128,128} 
\newcommand{\supp}{\textcolor{suppcolor}{\textit{supplementary material}}} 
\usepackage{xspace}
\def\eg{\emph{e.g}.,\xspace}

\def\ie{\emph{i.e}.,\xspace}

\usepackage{enumitem} 

\AtBeginDocument{%
  }

\renewcommand\footnotetextcopyrightpermission[1]{}
\begin{document}

\newcommand{\methodname}{RetroMem}
\title{Retrospective Memory for Camouflaged Object Detection}

\author{Chenxi Zhang}
\email{zhangchenxi2025@163.com}
\affiliation{%
  \institution{Shanghai Institute of Technology}
  \city{Shanghai}
  \country{China}
}
\author{Jiayun Wu}
\email{wujiayun_110@163.com}
\affiliation{%
  \institution{Shanghai Institute of Technology}
  \city{Shanghai}
  \country{China}
}

\author{Qing Zhang}
\authornote{Corresponding author}
\email{zhangqing@sit.edu.cn}
\affiliation{%
  \institution{Shanghai Institute of Technology}
  \city{Shanghai}
  \country{China}
}

\author{Yazhe Zhai}
\email{zhaiyazhe@njust.edu.cn}
\affiliation{%
  \institution{Nanjing University of Science}
  \city{Nanjing}
  \country{China}
}

\author{Youwei Pang}
\authornotemark[1]
\email{lartpang@gmail.com}
\affiliation{%
  \institution{Dalian University of Technology}
  \city{Dalian}
  \country{China}
}









\begin{abstract}
    Camouflaged object detection (COD) primarily focuses on learning 
    subtle yet discriminative representations from complex scenes. 
    Existing methods predominantly follow the parametric feedforward architecture based on static visual representation modeling. 
    However, they lack explicit mechanisms for acquiring historical context, limiting their adaptation and effectiveness in handling challenging camouflage scenes. 
    In this paper, we propose a recall-augmented COD architecture,  \ie \textbf{\methodname}, which  dynamically modulates camouflage pattern perception and inference by integrating relevant historical knowledge into the process. 
    Specifically, \methodname{} employs a two-stage training paradigm consisting of a learning stage and a recall stage to construct, update, and utilize memory representations effectively.
    During the learning stage, we design a dense multi-scale adapter (DMA) to improve the pretrained encoder's capability to capture rich multi-scale visual information with very few trainable parameters, thereby providing foundational inferences. 
    In the recall stage, we propose a dynamic memory mechanism (DMM) and an inference pattern reconstruction (IPR).
    These components fully leverage the latent relationships between learned knowledge and current sample context to reconstruct the  inference of camouflage patterns, thereby significantly improving the model's understanding of camouflage scenes. 
    Extensive experiments on several widely used datasets demonstrate that our \methodname{} significantly outperforms existing state-of-the-art methods.
\end{abstract}

\begin{teaserfigure}
\centering
  \includegraphics[width=0.99\textwidth]{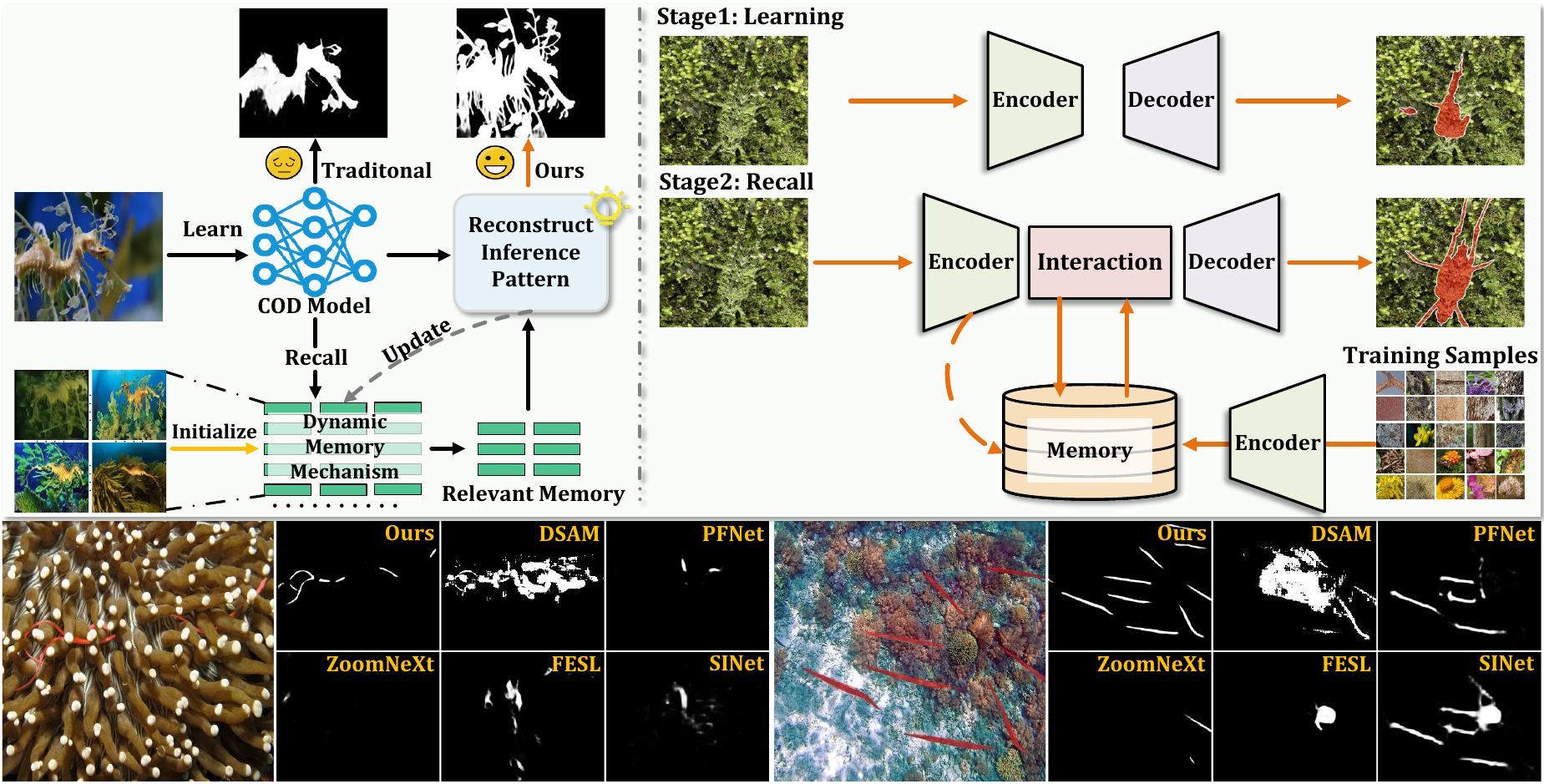}
  \caption{%
  \textit{Left}:
  Overview of the proposed method, featuring a dynamic memory mechanism for learned knowledge recall and memory representation updating to handle challenging camouflage cases more effectively.
  \textit{Right}:
  Two-stage pipeline of our method consisting of learning and recall stages.
  \textit{Bottom}:
  Visual comparison with SOTA methods in challenging camouflage scenes, \eg cluttered backgrounds (\textit{left}) and multiple objects (\textit{right}), showcasing our model's superior robustness.
  }
  \Description{}
  \label{fig:fig1}
\end{teaserfigure}


\maketitle

\section{Introduction}
\label{sec:introduction}

In nature, many animals adopt sophisticated camouflage patterns to interfere with predators' visual perception, thereby effectively concealing themselves within their environments.
The core of the camouflaged object detection (COD) task is to accurately segment these concealed objects from scenes that exhibit high visual similarity between targets and backgrounds. 
Nevertheless, the inherent diversity of camouflage patterns and the complexity of real-world scenes significantly increase the difficulty of this task. 
Due to its practical significance, COD research has been broadly applied across various domains, including 
wildlife protection~\cite{nafus2015hiding}, 
medical image analysis~\cite{fan2020pranet}, 
and agricultural monitoring~\cite{rustia2020application}.
Early COD methods rely on handcrafted features, such as texture~\cite{texture},
motion flow~\cite{flow}
and color contrast~\cite{color}.
These pioneering methods, however, suffer from inherent limitations in robustness and flexibility, restricting their applicability to diverse and complex visual scenes.
Recent advances in deep learning have substantially driven the COD field including
attention mechanisms~\cite{zhang2025rethinking,Camoformer,FSEL},
bio-inspired perception~\cite{ZoomNeXt,zoomnet2022},
and multi-task joint frameworks~\cite{BGNet,FEDER2023}
to extract subtle yet critical visual cues and construct multi-scale/view joint feature modeling for camouflage scene understanding.
Despite notable improvements, existing methods remain fundamentally constrained by their architectural paradigms.
The prevailing approach
excessively prioritizes the design of sequentially stacked modules for progressive feature refinement,
while adhering to static inference patterns reliant on instantaneous visual cues.
This dual focus however neglects the systematic integration of historical contextual knowledge and fails to enable effective reuse of acquired representational capabilities during network learning.
Consequently, these architectures exhibit critical deficiencies in dynamic adaptability and responsive mechanism design.
These constraints impair their performance to handle challenging camouflage scenes with diverse object characteristics and environmental conditions.
Targeted benchmark evaluations on unseen/rare scenes (Tab.~\ref{tab:seen}) reveal significant performance degradation in more challenging cases.
The results expose existing architectural limitations in scene adaptability.
These findings necessitate the COD framework design capable of synergizing historical contextual knowledge with adaptive inference pattern reconstruction.
This approach enables the context-aware inference mechanism to achieve robust cross-scene generalization.
Inspired by the above discussion, we propose \textbf{\methodname{}} (Fig.~\ref{fig:fig1} \textit{Left}), a novel COD paradigm inspired by the human brain's recall mechanism~\cite{Vilberg2008Memory}.
This framework simulates human-like adaptive inference through dynamically reconstructing inference pattern.
It injects historical knowledge into the current observation, which moves beyond static feature reliance to enhance generalization in complex camouflage scenes (Fig.~\ref{fig:fig1} \textit{Bottom}).
Grounded in the Confucian principle \textit{``Learn from the past and understand the present''} (Fig.~\ref{fig:fig1} \textit{Right}),
\methodname{} implements a \textbf{two-stage training framework} for strategic knowledge integration.
The initial \textbf{learning stage} deploys the dense multi-scale adapter (DMA) to finetune DINOv2~\cite{oquab2023dinov2}-based visual encoding.
This establishes foundational perception and segmentation capabilities for camouflaged objects.
The subsequent \textbf{recall stage} addresses architectural limitations by further introducing a learnable dynamic memory bank.
It stores and updates acquired knowledge during training, and enables adaptive recall of semantic-relevant representations during inference.
Considering knowledge transfer challenges from camouflage pattern variations between historical and current samples, we develop an inference pattern reconstruction (IPR).
It dynamically synthesizes memory-derived historical insights with current visual evidence, significantly boosting adaptive inference capacity.
Specifically, \methodname{} initially employs the segmentation capability acquired during the learning stage to identify subtle visual indicators suggestive of camouflaged objects.
It then recalls the most relevant prototype representations from the memory bank as historical context.
Finally, the IPR module integrates these components to reconstruct specific inference patterns for current samples.
Such a design improves the model performance in diverse camouflage challenges through context-aware decision-making.

Our main contributions are summarized as follows:
\begin{itemize}
    \item We propose \textbf{\methodname}, a two-stage training framework designed to reconstruct the architectural paradigm for COD, 
    mitigating the reliance on the instantaneous visual cues and the static inference pattern in conventional designs.
    \item We introduce the dense multi-scale adapter to effectively fine-tune the visual encoder for high-performance COD.
    \item We design a learnable dynamic memory mechanism (DMM) that efficiently stores and updates the contextual memory, enabling adaptive knowledge recall.
    \item We propose the inference pattern reconstruction to inject the recalled knowledge, which dynamically improves the sample-specific visual inference.
    \item Experiments on multiple conventional and carefully-constructed challenging COD benchmarks demonstrate the effectiveness and generalization of \methodname{}.
\end{itemize}

\begin{figure*}[!t]
    \centering
    \includegraphics[width=0.8\linewidth]{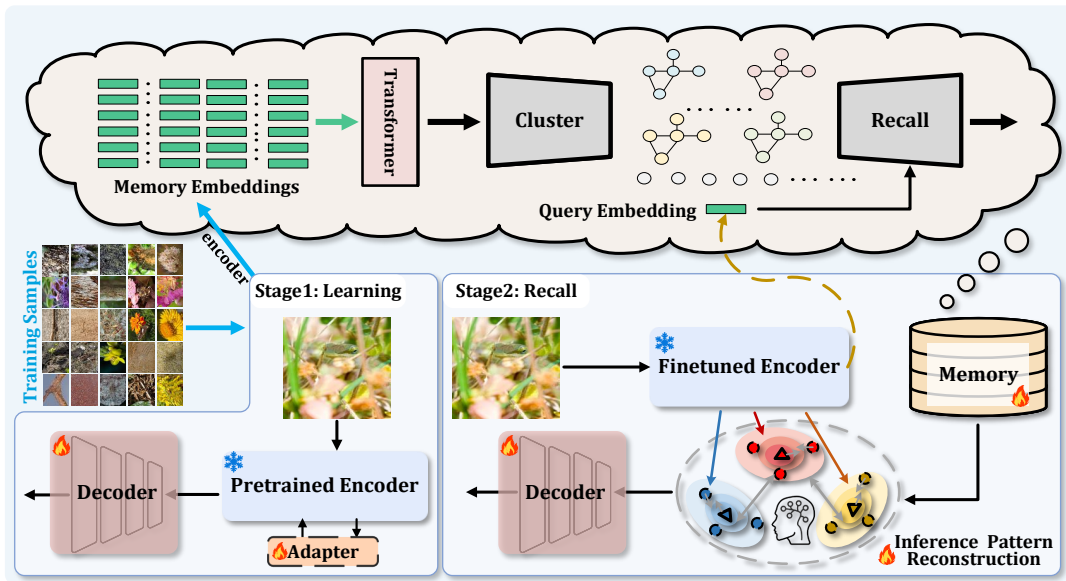}
    \caption{Overview of the proposed \methodname{} based on a two-stage training paradigm.
    The learning stage constructs a conventional COD network with multi-scale dense adapters to efficiently fine-tune the pretrained encoder, while embedding the training set into the memory.
    During the recall stage, specific inference patterns are dynamically reconstructed through recall of relevant memory embeddings for each input sample.
    }
    \label{fig:net}
\end{figure*}

\section{Related Work}
\label{sec:related_work}

\subsection{Camouflaged Object Detection (COD)}
\label{sec:related_cod}

Camouflage scenes are characterized by subtle visual cues that are frequently imperceptible to human observers.
However, these cues encapsulate critical contextual information exposing latent patterns in visual data.
The fundamental objective of COD resides in extracting these concealed yet discriminative features within intricate environments, thereby enabling accurate detection of camouflaged objects and recovery of their intrinsic structural characteristics.

Recent years have seen rapid progress in COD research, driven by novel approaches that enhance scene understanding through advanced perception mechanisms.
Biologically-inspired methods demonstrate particular efficacy.
The two-stage search-and-recognition processes are adopted in~\cite{SINet,PFNet} to first localize and then segment objects with increasing precision.
Some works~\cite{ZoomNeXt,zoomnet2022} employ dynamic multi-scale integration to resolve semantic relationships between ambiguous objects and their contexts.
Concurrently, multi-task paradigms improve object discernment by jointly optimizing auxiliary tasks like edge detection~\cite{BGNet,FSPNet,VSCode}.
Spectral processing techniques~\cite{FSEL,FEDER2023} are also introduced to integrate global representation and extract discriminative details in the frequency domain, effectively amplifying subtle camouflage patterns.
Additionally, several methods~\cite{Camoformer,zhang2025rethinking} explore intrinsic foreground-background interactions for more robust segmentation. 
Beyond the aforementioned RGB-centric methods, emerging COD methods explore cross-modal synergies through complementary semantic integration.
Representative implementations include geometric-aware frameworks~\cite{RISNet,DSAM} that leverage depth-derived structural priors to resolve spatial ambiguities, and vision-language paradigms~\cite{ACUMEN,zhang2024cgcod} that incorporate textual descriptors to establish semantic correspondences with camouflaged object representations.
Despite these advances, current COD methods remain hampered by their reliance on
the instantaneous visual cues and the sample-common inference mechanism,
struggling in complex scenes (Fig.~\ref{fig:fig1} \textit{Bottom}).
Like students excelling in drills but failing in practice, they perform well on test samples that mirror training data but falter under challenging scenes (\eg rare, unseen camouflage scenes as discussed in Sec.~\ref{sec:ablation_study} and Tab.~\ref{tab:seen}). 
This demands an adaptive mechanism that integrates learned knowledge with runtime contextual inference.

\subsection{Memory Mechanism}
\label{sec:related_memory}

The concept of memory is deeply rooted in human cognition, enabling individuals to store and recall past experiences to guide future perception and decision-making~\cite{lake2017building,hassabis2017neuroscience,barrett2019analyzing}. Inspired by this biological foundation, memory mechanisms have been incorporated into artificial intelligence systems, enabling models to retain and reuse informative representations over time. In deep learning, the Long Short-Term Memory (LSTM) network~\cite{hochreiter1997long} is among the earliest architectures to explicitly introduce memory into neural networks, effectively capturing long-range dependencies in sequential data. 
Building on this foundation, memory mechanisms have been widely adopted in tasks like video segmentation and anomaly detection, aiming to establish stable reference representations to enhance object region discrimination. In video segmentation, models such as XMem~\cite{cheng2022xmem} and RMem~\cite{oh2019video} mimic human sensory memory through multi-level memory hierarchies and attention-based retrieval, facilitating robust long-term tracking. Others, including CUTIE~\cite{yang2021cutie} and CompFeat~\cite{lu2020compfeat}, shift from pixel-level to object-level memory representations to achieve better generalization, while some approaches~\cite{voigtlaender2019feelvos} use dynamic memory updates to retain only essential information.
In anomaly detection, early methods based on autoencoders relied on reconstruction errors but often suffered from overfitting, limiting anomaly discrimination. MemAE~\cite{gong2019memae} addressed this by incorporating an external memory module with attention-based retrieval to reconstruct inputs using prototypical normal features. More recent patch-level methods, such as PatchCore~\cite{roth2022towards} and SoftPatch~\cite{jiang2022softpatch}, construct non-parametric memory banks from local features and employ KNN~\cite{cover1967nearest} for improved anomaly localization. TailedCore~\cite{jung2025tailedcore} further enhances robustness by selectively retaining rare-class features and applying noise filtering to achieve a balanced memory bank.
Unlike anomaly detection and video segmentation, which rely on stable reference patterns such as normality in appearance or temporal consistency across frames, 
COD is a \textit{class-agnostic} task in which the model is required to handle previously unseen object categories during inference. And camouflaged object lacks both clear semantic structure and salient visual cues. These characteristics limit the effectiveness of conventional memory-based approaches. 
To address these limitations, we propose a dynamic memory mechanism specifically tailored for COD. Instead of relying on fixed or temporally aligned references, our method constructs a memory bank from diverse training samples. During inference, the model recalls semantically similar memory based on input features, providing essential contextual cues to accurately identify camouflaged objects.

\section{Methodology}
\label{sec:methodology}

\subsection{Overall Architecture}
\label{sec:overall_architecture}

As depicted in Fig.~\ref{fig:net}, the proposed \methodname{} framework combines two synergistic stages.
The learning stage constructs a simple conventional COD model to encode all training samples into high-level features, forming an initialized memory bank.
During the recall stage, the framework dynamically recalls the most semantically relevant prototype from this bank for each input sample, aligns its visual features through the inference pattern reconstruction, and processes the reconstructed feature flow via the decoder to generate segmentation results.
Notably, the memory bank undergoes iterative improvement during recall-stage training by clustering semantically similar embeddings, which progressively enhances the stored representations to encapsulate discriminative knowledge.
This closed-loop architecture enables simultaneous feature preservation and adaptive inference pattern adjustment.

\subsection{Stage 1: Learning}
\label{sec:learning_stage}


In the learning stage, our COD model consists of a DINOv2 ~\cite{oquab2023dinov2}-based encoder and a simple multi-scale feature decoder.
Given an input image $X_l \in \mathbb{R}^{H \times W \times 3}$, we first extract multi-scale features $\{F_i\}_{i=1}^{3}$ from the encoder, and then progressively integrate them to obtain the camouflaged object prediction in the decoder.

\begin{figure}[!t]
    \centering
    \includegraphics[width=0.7\linewidth]{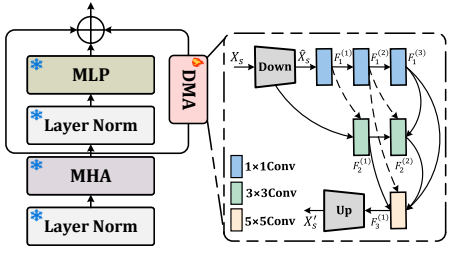}
    \caption{Illustration of the dense multi-scale adapter (DMA).}
    \label{fig:DMA}
\end{figure}

\noindent\textbf{Dense Multi-scale Adapter (DMA).}
%
%
Although DINOv2~\cite{oquab2023dinov2} shows strong transfer learning performance across various downstream vision tasks, the conventional full finetuning strategy may yield suboptimal results in the COD task, as shown in \supp{} (Sec. B.4). 
This stems from the tension between the massive trainable parameters in DINOv2 and the relatively limited training data in COD, creating high risks of overfitting.
Furthermore, while multi-scale information integration is known to be critical for COD~\cite{zoomnet2022,ZoomNeXt}, naively finetuning all layers fails to explicitly exploit such inductive biases.
To this end, we propose the DMA shown in Fig.~\ref{fig:DMA} to transform the pretrained DINOv2 into a COD specialist via lightweight architectural adaptation.
By injecting trainable dense fusion paths in the frozen backbone, the DMA preserves DINOv2's generalizable visual representations by freezing original parameters and enables context-aware aggregation of cross-scale features for precise camouflage object perception.
%
%
Specifically, our DMA first utilizes a linear projection layer (${Linear}_{\text{down}}$) to reduce the dimension of the input feature $X_s$:
\begin{equation}
    \hat{X}_s =\sigma({Linear}_{\text{down}}({X}_s))
\end{equation}
where $\sigma$ is the non-linear activation function, \ie ReLU.
Then, we use three $1\times1$ convolutional layers to generate features $\{F_1^{(n)}\}_{n=1}^3$: 
\begin{equation}
    F_1^{(n)} = 
    \begin{cases}
        {Conv}_{1\times1}(\hat{X}_s)   & \text{if } n=1 \\
        {Conv}_{1\times1}(F_1^{(n-1)}) & \text{if } n=2,3
    \end{cases}
\end{equation}
where ${Conv}_{k\times{k}}$ represents the convolution layer with a kernel size of $k$.
%
Next, mid-scale context representation $\{F_2^{(n)}\}_{n=1}^2$ is extracted by $3 \times 3$ convolution layers and integrated with $F_1^{(n)}$ to enhance structural details:
\begin{align}
    F_2^{(1)} & = {Conv}_{3\times3} \left( {Conv}_{1\times1} ([\hat{X}_s , F_1^{(1)}]) \right)                 \\
    F_2^{(2)} & = {Conv}_{3\times3} \left( {Conv}_{1\times1} ([F_2^{(1)} , F_1^{(2)} , F_1^{(3)}]) \right)
\end{align}
where $[\cdot]$ denotes the channel-wise concatenation operation.
Additionally, a $5 \times 5$ convolutional layer aggregates these features to capture information across different receptive fields, thus effectively discriminating local details:
\begin{equation}
    F_3^{(1)} = {Conv}_{5\times5} \left( {Conv}_{1\times1} ([F_2^{(1)} , F_2^{(2)} , F_1^{(2)} , F_1^{(3)}]) \right)
\end{equation}
The final output $X'_s$ of the DMA is obtained via a linear up-projection layer ${Linear}_{\text{up}}$ which restores the original channel dimension: 
\begin{equation}
    {X'}_s = {Linear}_{\text{up}}(F_3^{(1)})
\end{equation}
%
Instead of merely combining multi-scale features, our DMA hierarchically aggregates localized patterns across varying scales through dense connections, ensuring progressive refinement while preserving fine-grained details.

\subsection{Stage 2: Recall}
\label{sec:recall}

This stage aims to alleviate the generalization bottleneck of the model through the learnable dynamic memory coupled with adaptive inference pattern reconstruction, enabling dual adaptation to diverse camouflage scenes.
Initially, feature embeddings from training samples are adaptively clustered into a prototype memory bank $\mathcal{M}$ that encodes representative prototypes of various camouflage patterns.
Multi-level features $\{F_i\}^3_{i=1}$ extracted by the frozen encoder drive cosine similarity-based retrieval of the most relevant prototype $\hat{p}_q$ from $\mathcal{M}$, with the deepest feature $F_3$ serving as the query embedding.
This retrieved representation triggers sample-specific pattern reconstruction through semantic attribute clustering to dynamically align the historical context with current features.
The mechanism preserves fidelity to learned camouflage patterns via memory-based recall while achieving the adaptation to diverse scenes through the online-updated inference pattern.

\noindent\textbf{{Dynamic Memory Mechanism (DMM).}}
The memory bank $\mathcal{M}$ is initialized by processing all $N$ training images through the Stage~1-finetuned encoder, where $\ell_2$-normalized and globally average-pooled features form preliminary embeddings $M \in \mathbb{R}^{N \times C}$.
These embeddings are enhanced as $M_{e} \in \mathbb{R}^{N \times C}$ through a single transformer layer~\cite{MHSA}.
$M_{e}$ is subsequently assigned to their respective clusters and converted to $M_c \in \mathbb{R}^{N \times C}$ by HDBSCAN~\cite{mcinnes2017hdbscan}, which adaptively groups similar embeddings into $K$%
\footnote{Due to the nature of HDBSCAN, $K$ does not need to be specified manually.}
semantically distinctive clusters with center prototypes \( P_c = \{p_k\}_{k=1}^{K} \).
This process not only mitigates computational redundancy and noisy interferences from raw sample-level storage and recall, but also captures shared semantic characteristics among different camouflage instances.
During training, we iteratively optimize the transformer parameters and update the memory bank by constraining the consistency of the representation $M_{e}$ with the clustered memory prototypes as detailed in Sec.~\ref{sec:loss_function}.
By this way, the memory bank can be refined into a compact and discriminative form.
%
%
Next, the cosine distance is used to determine the sample-prototype similarity for the query embedding \( f_q \) from the deepest layer in the encoder:
\begin{equation}
    s_{q,k} = \frac{\langle f_q, p_k \rangle}{\|f_q\|_2 \|p_k\|_2}, \quad k = 1, 2, \dots, K
\end{equation}
where \( \langle \cdot, \cdot \rangle \) and \( \|\cdot\|_2 \) are the inner product between vectors and the \( \ell_2 \) norm, respectively.
And the most relevant prototype is:
\begin{equation}
    \hat{p}_q = {P}_c[\arg\max_k s_{q,k}] \in {R}^{C}
\end{equation}
The final prototype $\hat{p}_q$ is used to adjust and reconstruct the forward propagation of the multi-level encoder features.
%

\begin{figure}[!t]
    \centering
    \includegraphics[width=1\linewidth]{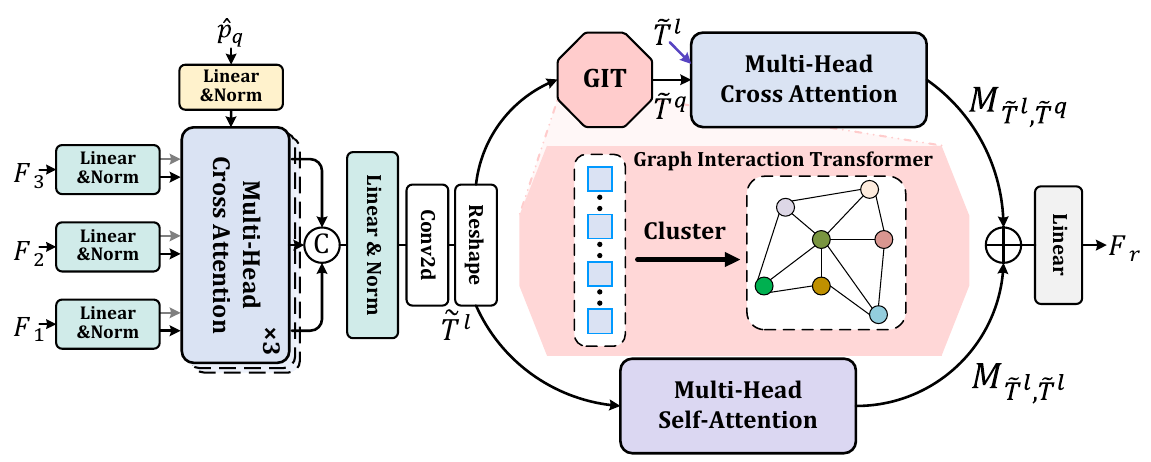}
    \caption{Illustration of the inference pattern reconstruction.}
    \label{fig:RPR}
\end{figure}

\noindent\textbf{Inference Pattern Reconstruction (IPR).}
The recalled memory prototype $\hat{p}_q$ serves as a condensed and semantically rich representation of a cluster of camouflage patterns, offering valuable inference context.
So, we introduce the IPR shown in Fig.~\ref{fig:RPR} to integrate $\hat{p}_q$ in a context-aware manner.
By establishing multi-dimensional correlations across these heterogeneous feature sources, it enables robust and adaptive inference across diverse camouflage scenes. 
%
The IPR employs a multi-step feature interaction strategy to integrate the memory embedding $\hat{p}_q$ and multi-level visual features at multiple scales, enhancing hidden correlations in latent dimensions. 
We first introduce multi-head cross-attention~\cite{MHSA} to compute their correlations. 
Given the embedding \( \hat{p}_q \) and multi-level encoder features \( \{F_i\}_{i=1}^{3} \), the query $F_i^q$, key $T^k$ and value $T^v$ are computed by the projection matrices $W_q$, $W_k$ and $W_v$:
\begin{align}
    F_i^q &= \operatorname{reshape}(F_i) W_q \in \mathbb{R}^{H_iW_i \times C^l}, \quad i \in \{1, 2, 3\}\\
    T^k &= \operatorname{reshape}(\hat{p}_q) W_k, T^v = \operatorname{reshape}(\hat{p}_q) W_v \in \mathbb{R}^{1 \times C^l}, \\
    T_i^l &= \operatorname{Softmax}\left( \frac{F_i^q (T^k)^\top}{\sqrt{C^l}} \right) T^v \in \mathbb{R}^{H_iW_i \times C^l}
\end{align}
where $C^l$ is the channel number of $F_i^q$, $T^k$ and $T^v$.
%
%
To further enhance semantic expression, the refined feature \( \tilde{T}^l \) is first projected into a latent graph space, where similar semantic features are clustered into graph nodes.
And then, we introduce the graph interaction transformer (GIT)~\cite{yao2024hierarchical} in IPR to aggregate these features as $\tilde{T}^q$, which serves as a structured representation capturing long-range contextual information.
By leveraging graph-based clustering, the GIT ensures that similar camouflage patterns are represented in a compact and consistent manner, facilitating robust inference across complex scenes.
Then, the cross-attention between \( \tilde{T}^q \) and \( \tilde{T}^l \) and the self-attention of \( \tilde{T}^l \) aggregate and reconstruct the inference pattern, yielding the final enhanced feature $F_r$:
\begin{align}
    M_{\tilde{T}^l \leftarrow \tilde{T}^o} & = \operatorname{Softmax}\left( \frac{W_q \tilde{T}^o (W_k \tilde{T}^l)^\top}{\sqrt{C^l}} \right) W_v \tilde{T}^l \\
    F_r & =  M_{\tilde{T}^l , \tilde{T}^l} + M_{\tilde{T}^l , \tilde{T}^q}
\end{align}
where $\tilde{T}^o$ denotes either $\tilde{T}^q$ or $\tilde{T}^l$.
The output feature $F_r$ serves as guidance for subsequent inference. 
%
%
Rather than relying solely on the feature similarity, our IPR adaptively models contextual consistency and constructs the sample-specific inference process by integrating the current representation with the semantic-relevant memory prototype.
It enables the model generalization across diverse camouflage scenes by aligning and combining historical knowledge with the current input, thereby enhancing accuracy and robustness. 
%

\subsection{Decoder Network}
\label{sec:decoder}

To effectively utilize the inference guidance ${F}_r$, we design a simple yet efficient decoder.
We employ the convolutional LSTM ($\Gamma$) to progressively fuse multi-level features $\{F_i\}_{i=1}^{3}$, where ${F}_r$ serves as a long-term memory item:
\begin{equation}
    \tilde{F}_i =
    \begin{cases}
             
        {\Gamma}(F_3, F_3)+F_3 & \text{if } i=3 \\
        {\Gamma}(F_i, [\operatorname{Up}(\tilde{F}_{i+1}), \operatorname{Up}(F_3)])+\operatorname{Up}(\tilde{F}_{i+1}) & \text{if } i=2,1 \text{ in Sec.~\ref{sec:learning_stage}} \\
        {\Gamma}(F_i, [\operatorname{Up}(\tilde{F}_{i+1}), \operatorname{Up}(F_r)])+\operatorname{Up}(\tilde{F}_{i+1}) & \text{if } i=2,1 \text{ in Sec.~\ref{sec:recall}} 
    \end{cases}
\end{equation}
where ``$\operatorname{Up}$'' is the bilinear interpolation for spatial alignment.
Finally, the $1 \times 1$ convolution is applied to reduce the channel dimension and produce the segmentation predictions $\{P_i\}_{i=1}^{3}$ from features $\{\tilde{F_i}\}_{i=1}^{3}$, respectively.

\subsection{Loss Function}
\label{sec:loss_function}

Following existing COD methods~\cite{ZoomNeXt,RISNet}, we adopt a hybrid loss ($\mathcal{L}_{\text{seg}} = \mathcal{L}_{{\text{bce}}} + \mathcal{L}_{{\text{iou}}}$) consisting of the binary cross-entropy loss ($\mathcal{L}_{{\text{bce}}}$) and the IoU loss ($\mathcal{L}_{{\text{iou}}}$) during training.
This hybrid loss provides a good balance between pixel-wise accuracy and region-level consistency.
Besides, we introduce a consistency loss $\mathcal{L}_c$, which encourages the transformer in our memory to form structurally and semantically consistent feature spaces with the clustering results by minimizing the distance between the each instant $M_e[i] \in \mathbb{R}^{C}$ and its corresponding cluster prototype $M_c[i] \in \mathbb{R}^{C}$:
\begin{align}
    \mathcal{L}_c(M_e, M_c) = \frac{1}{N} \sum_{i=1}^{N} \|M_e[i] - M_c[i]\|_2^2
\end{align}
where the embeddings $M_e, M_c \in \mathbb{R}^{N \times C}$ are from the transformer layer and the clustering algorithm as described in Sec.~\ref{sec:recall}, respectively.
%
So, our overall loss function is defined as:
\begin{align}
    \mathcal{L} & = \sum_{i=1}^{3} \mathcal{L}_{\text{seg}}(P_i, G) + \mathcal{L}_c(M_e, M_c)
\end{align}
where $\{P_{3}, P_{2}, P_{1}\}$ represent the predictions generated by our decoder and \({G}\) refers to the ground truth mask.


\begin{table*}[!t]
    \centering
     \caption{Quantitative results.
    Top-three results are highlighted in \textcolor{red}{\textbf{red}}, \textcolor{green}{\textbf{green}}, and \textcolor{blue}{\textbf{blue}}.
    ``-'' indicates publicly unavailable results.}
    \resizebox{0.9\linewidth}{!}{

\begin{tabular}{l|c|c|ccccc|ccccc|ccccc}
\toprule
\multirow{2}{*}{Methods} &
\multirow{2}{*}{Year} &
\multirow{2}{*}{Backbone} &
\multicolumn{5}{c|}{CAMO (250 images)} &
\multicolumn{5}{c|}{COD10K (2,026 images)} &
\multicolumn{5}{c}{NC4K (4,121 images)} \\
& & & $S_{\alpha}\uparrow$ & $F^{\omega}_{\beta}\uparrow$ & $F_{\beta}\uparrow$ & ${M}\downarrow$ & $E_m\uparrow$ 
& $S_{\alpha}\uparrow$  & $F^{\omega}_{\beta}\uparrow$ & $F_{\beta}\uparrow$ & ${M}\downarrow$ & $E_m\uparrow$ 
& $S_{\alpha}\uparrow$  & $F^{\omega}_{\beta}\uparrow$ & $F_{\beta}\uparrow$ & ${M}\downarrow$ & $E_m\uparrow$ \\

\midrule
\multicolumn{18}{c}{\centering \textbf{CNN and Transformer based Methods}} \\
\midrule

SINet~\cite{SINet} & CVPR'2020 & ResNet~\cite{residual2016} & 0.751 & 0.644 & 0.702 & 0.100 & 0.771 & 0.751 & 0.551 & 0.679 & 0.051 & 0.806 & 0.808 & 0.723 & 0.769 & 0.058 & 0.887 \\
PFNet~\cite{PFNet} & CVPR'2021 & ResNet~\cite{residual2016} & 0.782 & 0.695 & 0.746 & 0.085 & 0.855 & 0.800 & 0.660 & 0.701 & 0.040 & 0.877 & 0.829 & 0.745 & 0.784 & 0.053 & 0.887 \\
BGNet~\cite{BGNet} & IJCAI'2022 & Res2Net~\cite{Res2Net} & 0.812 & 0.749 & 0.789 & 0.073 & 0.870 & 0.831 & 0.722 & 0.753 & 0.033 & 0.901 & 0.851 & 0.788 & 0.820 & 0.044 & 0.907 \\
ZoomNet~\cite{zoomnet2022} & CVPR'2022 & ResNet~\cite{residual2016} & 0.820 & 0.752 & 0.794 & 0.066 & 0.877 & 0.838 & 0.729 & 0.766 & 0.029 & 0.888 & 0.853 & 0.784 & 0.818 & 0.043 & 0.896 \\
FEDER~\cite{FEDER2023} & CVPR'2023 & Res2Net~\cite{Res2Net} & 0.836 & 0.807 & 0.781 & 0.066 & 0.897 & 0.844 & 0.748 & 0.751 & 0.029 & 0.911 & 0.862 & 0.824 & 0.824 & 0.042 & 0.913 \\
FSPNet~\cite{FSPNet} & CVPR'2023 & ViT~\cite{ViT} & 0.851 & 0.802 & 0.831 & 0.056 & 0.905 & 0.850 & 0.755 & 0.769 & 0.028 & 0.912 & 0.879 & 0.816 & 0.843 & 0.035 & 0.914 \\
HitNet~\cite{HitNet} & AAAI'2023 & PVT-B2~\cite{pvt} & 0.849 & 0.809 & 0.831 & 0.055 & 0.906 & 0.871 & 0.806 & 0.823 & 0.023 & 0.935 & 0.875 & 0.834 & 0.854 & 0.037 & 0.926 \\
PRNet~\cite{PRNet} & TCSVT'2024 & PVT-B2~\cite{pvt} & 0.872 & 0.831 & 0.855 & 0.050 & 0.922 & 0.874 & 0.799 & 855 & 0.023 & 0.937 & 0.891 & 0.848 & 0.869 & 0.031 & 0.935 \\
DSAM~\cite{DSAM} & MM'2024 & SAM~\cite{chen2023sam} & 0.825 & 0.798 & 0.824 & 0.061 & 0.917 & 0.837 & 0.766 & 0.789 & 0.033 & 0.925 & 0.864 & 0.832 & 0.851 & 0.040 & 0.932 \\
FSEL~\cite{FSEL} & ECCV'2024 & PVT-B4~\cite{pvt} & 0.885 & 0.857 & 0.868 & 0.040 & 0.942 & 0.877 & 0.799 & 0.817 & 0.021 & 0.937 & 0.892 & 0.852 & 0.871 & 0.030 & 0.941 \\
ACUMEN~\cite{ACUMEN} & ECCV'2024 & CLIP-L~\cite{radford2021learning} & 0.886 & 0.850 & 0.861 & 0.039 & 0.939 & 0.852 & 0.761 & 0.778 & 0.026 & 0.930 & 0.874 & 0.826 & 0.844 & 0.036 & 0.932 \\
CamoFormer~\cite{Camoformer} & TPAMI'2024 & PVT-B4~\cite{pvt} & 0.872 & 0.831 & 0.854 & 0.046 & 0.929 & 0.869 & 0.786 & 0.842 & 0.023 & 0.932 & 0.892 & 0.847 & 0.868 & 0.030 & 0.939 \\
ZoomNeXt~\cite{ZoomNeXt} & TPAMI'2024 & PVT-B4~\cite{pvt} & 0.888 & 0.859 & 0.878 & 0.040 & 0.943 & 0.898 & 0.838 & 0.848 & 0.017 & 0.955 & \textcolor{blue}{\textbf{0.900}} & 0.865 & 0.884 & 0.028 & 0.949 \\

RISNet~\cite{RISNet} & CVPR'2024 & PVT-B2~\cite{pvt} & 0.870 & 0.827 & 0.844 & 0.050 & 0.922 & 0.873 & 0.799 & 0.817 & 0.025 & 0.931 & 0.882 & 0.834 & 0.854 & 0.037 & 0.926 \\
VSCode~\cite{VSCode} & CVPR'2024 & Swin-T~\cite{liu2021swin} & 0.873 & 0.820 & 0.844 & 0.046 & 0.925 & 0.869 & 0.780 & 0.806 & 0.025 & 0.931 & 0.882 & 0.841 & 0.863 & 0.032 & 0.935 \\
HGINet~\cite{yao2024hierarchical} & TIP'2024 & ViT+RTFA~\cite{ViT} & 0.874 & 0.848 & \textcolor{blue}{\textbf{0.868}} & 0.041 & 0.937 & 0.882 & 0.821 & 0.835 & 0.019 & 0.949 & 0.894 & 0.865 & 0.880 & 0.027 & 0.947 \\
CGNet~\cite{zhang2024cgcod} & Arxiv'2024 & CLIP~\cite{radford2021learning} & \textcolor{blue}{\textbf{0.896}} & 0.864 & \textcolor{green}{\textbf{0.882}} & \textcolor{blue}{\textbf{0.036}} & \textcolor{blue}{\textbf{0.947}} & 0.890 & 0.824 & 0.859 & 0.018 & 0.948 & 0.904 & 0.869 & 0.887 & \textcolor{blue}{\textbf{0.026}} & 0.949 \\
BiRefNet~\cite{zheng2024birefnet} & AIR'2024 & Swin-L~\cite{liu2021swin} & \textcolor{green}{\textbf{0.904}} & \textcolor{green}{\textbf{0.890}} & - & \textcolor{green}{\textbf{0.030}} & \textcolor{green}{\textbf{0.954}} & \textcolor{green}{\textbf{0.913}} & \textcolor{red}{\textbf{0.874}} & - & \textcolor{green}{\textbf{0.014}} & \textcolor{green}{\textbf{0.960}} & \textcolor{green}{\textbf{0.914}} & 0.894 & - & \textcolor{green}{\textbf{0.023}} & \textcolor{blue}{\textbf{0.953}} \\
CamoDiffusion~\cite{sun2025conditional} & TPAMI'2025 & PVT-B4~\cite{pvt} & 0.878 & 0.853 & - & 0.042 & 0.940 & 0.881 & 0.814 & - & 0.026 & 0.895 & 0.893 & 0.859 & - & 0.035 & 0.942 \\
\midrule
\multicolumn{18}{c}{\centering \textbf{ Finetuning-based Methods}} \\
\midrule
SAM-Adapter~\cite{chen2023sam} & ICCVW'2023 & SAM-H~\cite{chen2023sam} & 0.847 & 0.765 & - & 0.070 & 0.873 & 0.883 & 0.824 & - & 0.025 & 0.918 & - & - & - & - & - \\
FGSA-Net~\cite{FGSANet} & TMM'2024 & ViT-L~\cite{ViT} & 0.889 & \textcolor{blue}{\textbf{0.870}} & - & \textcolor{blue}{\textbf{0.036}} & 0.944 & 0.893 & 0.849 & - & \textcolor{blue}{\textbf{0.015}} & 0.953 & 0.903 & \textcolor{blue}{\textbf{0.883}} & - & \textcolor{green}{\textbf{0.023}} & 0.951 \\
SAM2-UNet~\cite{xiong2024sam2} & Arxiv'2024 & SAM2-L~\cite{kirillov2024sam2} & 0.884 & 0.844 & 0.864 & 0.042 & 0.932 & 0.880 & 0.789 & 0.800 & 0.021 & 0.936 & 0.901 & 0.856 & 0.872 & 0.029 & 0.941 \\
SAM2-Adapter~\cite{chen2024sam2} & Arxiv'2024 & SAM2-L~\cite{kirillov2024sam2} & 0.855 & 0.810 & - & 0.051 & 0.909 & 0.899 & 0.850 & - & 0.018 & 0.950 & - & - & - & - & - \\
\midrule
\methodname~ & Ours & DINOv2-B~\cite{oquab2023dinov2} & \textcolor{red}{\textbf{0.913}} & \textcolor{red}{\textbf{0.894}} &\textcolor{red}{\textbf{0.903}}& \textcolor{red}{\textbf{0.028}} & \textcolor{red}{\textbf{0.959}} & \textcolor{red}{\textbf{0.915}} & \textcolor{green}{\textbf{0.868}}&\textcolor{red}{\textbf{0.876}} & \textcolor{red}{\textbf{0.014}} & \textcolor{red}{\textbf{0.966}} & \textcolor{red}{\textbf{0.923}} & \textcolor{red}{\textbf{0.900}} &\textcolor{red}{\textbf{0.910}}& \textcolor{red}{\textbf{0.020}} & \textcolor{red}{\textbf{0.963}} \\
\bottomrule
\end{tabular}

}
    \label{tab:cod_comparison}
\end{table*}

\begin{table}[!t]
    \centering
    \caption{Performance on CHAMELEON~\cite{CHAMELEON} dataset.}
    \resizebox{0.9\linewidth}{!}{
\begin{tabular}{l|c|ccccc}
\midrule
\multirow{2}{*}{{Methods}} & \multirow{2}{*}{{Year}} & \multicolumn{5}{c}{CHAMELEON (76 images)} \\
& & $S_{\alpha}\uparrow$  & $F^{\omega}_{\beta}\uparrow$ & $F_{\beta} \uparrow$ & ${M}\downarrow$ & $E_m\uparrow$ \\
\midrule
SINet~\cite{SINet}         & CVPR'2020   & 0.869 & 0.740 & 0.790 & 0.044 & 0.899 \\
PFNet~\cite{PFNet}         & CVPR'2021   & 0.882 & 0.810 & 0.828 & 0.033 & 0.942 \\
BGNet~\cite{BGNet}         & IJCAI'2022  & 0.901 & 0.851 & 0.860 & 0.027 & 0.943 \\
ZoomNet~\cite{zoomnet2022} & CVPR'2022   & 0.902 & 0.845 & 0.864 & 0.023 & 0.952 \\
FEDER~\cite{FEDER2023}     & CVPR'2023   & 0.887 & 0.834 & 0.851 & 0.030 & 0.946 \\
FSPNet~\cite{FSPNet}       & CVPR'2023   & 0.908 & 0.851 & 0.867 & 0.022 & 0.943 \\
HitNet~\cite{HitNet}       & AAAI'2023   & 0.921 & 0.897 & 0.902 & 0.019 & 0.967 \\
PRNet~\cite{PRNet}         & TCSVT'2024  & 0.914 & 0.874 & 0.886 & 0.020 & 0.971 \\
FSEL~\cite{FSEL}           & ECCV'2024   & 0.916 & 0.880 & 0.889 & 0.022 & 0.958 \\
CamoFormer~\cite{Camoformer} & TPAMI'2024 & 0.910 & 0.865 & 0.882 & 0.022 & 0.957 \\
ZoomNeXt~\cite{ZoomNeXt}   & TPAMI'2024  & 0.924 & 0.885 & 0.896 & 0.018 & 0.975 \\
\methodname~ (Ours)        & -         & \textcolor{red}{\textbf{0.931}} & \textcolor{red}{\textbf{0.903}} & \textcolor{red}{\textbf{0.908}} & \textcolor{red}{\textbf{0.017}} & \textcolor{red}{\textbf{0.972}} \\
\bottomrule
\end{tabular}
\label{tab:chameleon}

}
    \label{tab:chameleon}
\end{table}

\section{Experiments}
\label{sec:experiments}

\subsection{Experimental Settings}
\label{sec:impplementation_details}

\noindent\textbf{Datasets.}
We train and evaluate the proposed \methodname{} on four widely used COD datasets.
The training set comprises samples from CAMO~\cite{CAMO} and COD10K~\cite{SINetV2}, totaling 4,040 images.
For evaluation, we utilize the CAMO, COD10K, NC4K~\cite{NC4K} and  CHAMELEON~\cite{CHAMELEON} datasets, containing 250, 2,026, 4,121, and 76 images, respectively.
More details can be found in the \supp{} (Sec. A.1). 

\noindent\textbf{Criteria.}
To assess model effectiveness, five commonly-used metrics are introduced:
{S-measure} (\(S_{\alpha}\))~\cite{Smeasure}, 
{E-measure} (\(E_m\))~\cite{Emeasure}, 
{weighted F-measure} (\(F_{\beta}^{\omega}\))~\cite{wf}, 
{adaptive F-measure} (\(F_{\beta}^{}\))~\cite{salvador2011superpixel}, 
and {mean absolute error} (\(M\)).
Higher values for \(S_{\alpha}\) , \(F_{\beta}^{\omega}\), \(F_{\beta}^{}\) and \(E_m\) indicate better performance, whereas a lower \(M\) signifies improved accuracy. 
More details about metrics can be found in the \supp{}.

\noindent\textbf{Implementation.}
The proposed \methodname{} is implemented using the PyTorch framework and trained on an NVIDIA RTX 3090 GPU.
All experiments adopt a batch size of 12.
Input images are resized to $448 \times 448$.
We utilize the AdamW~\cite{kingma2014adam} optimizer and employ the DINOv2-B/14~\cite{oquab2023dinov2} as the vision backbone.
The proposed dense multi-scale adapter (DMA) is applied at layers $\{1, 3, 5, 7, 9, 11\}$ within the frozen DINOv2. For layer selection, We performed an ablation study on layer selection in the \supp{}(Section. B.2 ) to validate the chosen DMA insertion layers.

\subsection{Comparison with State-of-the-Art Methods}

To evaluate the method effectiveness, we compare \methodname{} against several state-of-the-art conventional~\cite{SINet,PFNet,BGNet,zoomnet2022,FEDER2023,FSPNet,HitNet,PRNet,FSEL,PNet,Camoformer,ZoomNeXt,ACUMEN,VSCode,zheng2024birefnet,sun2025conditional} and finetuning-based~\cite{chen2023sam,FGSANet,xiong2024sam2, chen2024sam2} methods, as well as existing RGB-D~\cite{DSAM,RISNet} and vision-text methods~\cite{ACUMEN,zhang2024cgcod}.
For a fair comparison, all prediction results are obtained directly from the authors' official implementations.

\begin{figure}[!t]
    \centering
    \includegraphics[width=1\linewidth]{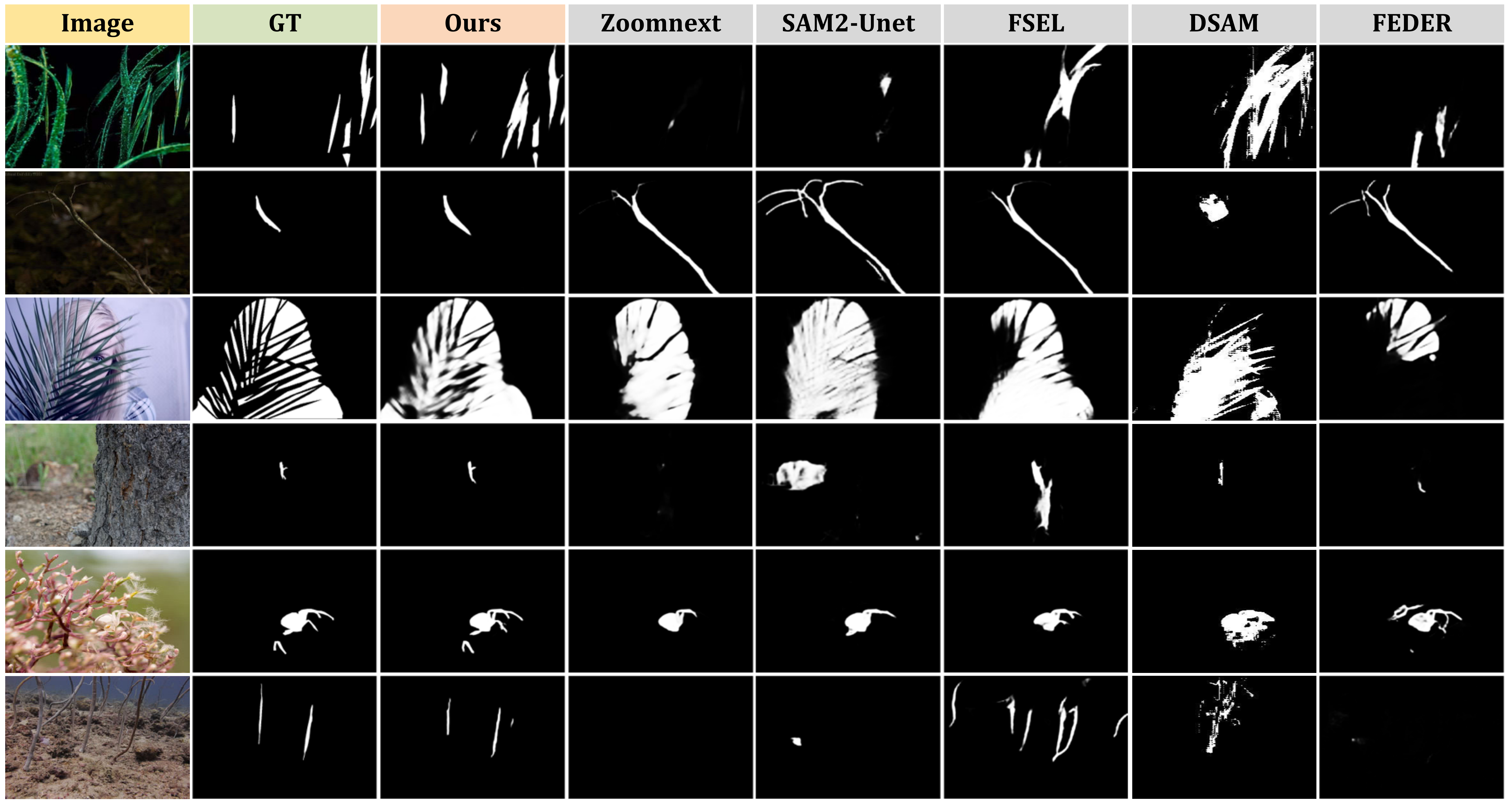}
    \caption{Representative visual results from various methods.}
    \label{fig:visual}
\end{figure}

\noindent\textbf{Quantitative Result.}
As shown in Tab.~\ref{tab:cod_comparison} and Tab.~\ref{tab:chameleon}, our \methodname{} outperforms all 23 competing methods, except for BiRefNet in $F^{\omega}_{\beta}$ on COD10K. 
Notably, compared to recent COD methods such as BiRefNet, RISNet, ACUMEN, and DSAM, which leverage high-resolution inputs or incorporate additional modalities information (\eg textual descriptions, depth maps) to enhance performance, our method exhibits strong performance across all benchmark datasets.
Specifically, compared to the high-resolution method BiRefNet, our model achieves average improvements of 1.2\%, 2.8\%, 1.4\%, and 15.1\% in $S_m$, $F^{\omega}_{\beta}$, $E_m$, and $M$, respectively.
Furthermore, although certain finetuning-based methods utilize more complex pre-trained backbones (\eg ViT-L, ViT-H) to enhance performance, a performance gap still persists in comparison with our approach.
In addition, \methodname{} surpasses the finetuning-based FGSA-Net~\cite{FGSANet} by 2.5\%, 7.9\%, 4.6\%, and 18.7\% in the same metrics. 

\noindent\textbf{Qualitative Result.}
As illustrated in Fig.~\ref{fig:c_pattern}, most camouflage patterns in the training dataset focus on background matching (\eg color, shape, texture), while rare cases such as extreme sizes, edge blurring, noise interference, multiple objects, and improper exposure are underrepresented.
Thus, we emphasize these challenging scenes in our evaluation. 
Fig.~\ref{fig:visual} compares our method with SOTA approaches.
For large and small objects ($3^{rd}$ and $4^{th}$ rows), \methodname{} achieves more complete segmentation while preserving contours.
In multi-object cases ($1^{st}$ and $6^{th}$ rows), it accurately detects all objects.
Under low-light conditions with noise interference (second row), our memory mechanism recalls relevant camouflage pattern knowledge, enabling correct segmentation where other methods fail. These results highlight the superior generalization capability of \methodname{} in complex camouflage scenes. We provide more visual comparisons in the \supp{} (Sec. C).

\subsection{Ablation Study}
\label{sec:ablation_study}

\begin{table}[!t]
    \centering
    \caption{Ablation comparison of proposed components. ``Param.'' means the number of trainable parameters.}
    \resizebox{\linewidth}{!}{

\begin{tabular}{c|l|ccc|ccccc|ccccc}
\toprule
\multirow{2}{*}{No.} & \multirow{2}{*}{Variants} & Param. & MACs & FPS 
& \multicolumn{5}{c|}{COD10K (2,026 images)} 
& \multicolumn{5}{c}{NC4K (4,121 images)} \\
& & (M) & (G) & 
&$S_{\alpha}\uparrow$  & $F^{\omega}_{\beta} \uparrow$ & $F_{\beta} \uparrow$ & $\mathcal{M} \downarrow$ & $E_m \uparrow$
& $S_{\alpha}\uparrow$  & $F^{\omega}_{\beta} \uparrow$ & $F_{\beta} \uparrow$ & $\mathcal{M} \downarrow$ & $E_m \uparrow$ \\
\midrule
\multicolumn{15}{r}{\textit{Stage 1: Learning}} \\
\midrule
I   & Baseline       & 0.53  & 27.90 & 71.81 
& 0.780 & 0.693 & 0.702 & 0.054 & 0.810 & 0.722 & 0.660 & 0.675 & 0.065 & 0.761 \\

II  & No. I+DMA      & 4.98  & 28.32 & 62.50
    & 0.881 & 0.833 & 0.844 & 0.023 & 0.930 
    & 0.898 & 0.846 & 0.860 & 0.030 & 0.941 \\
\midrule
\multicolumn{15}{r}{\textit{Stage 2: Recall}} \\
\midrule

III  & No. II+IPR     & 36.03 & 38.12 & 28.89
    & 0.908 & 0.855 & 0.864 & 0.017 & 0.948 
    & 0.914 & 0.869 & 0.864 & 0.025 & 0.953 \\

IV   & No. II+IPR+DMM & 40.01 & 39.84 & 22.68
    & \textbf{\textcolor{red}{0.915}} & \textbf{\textcolor{red}{0.868}} & \textbf{\textcolor{red}{0.886}} & \textbf{\textcolor{red}{0.014}} & \textbf{\textcolor{red}{0.966}} 
    & \textbf{\textcolor{red}{0.923}} & \textbf{\textcolor{red}{0.900}} & \textbf{\textcolor{red}{0.910}} & \textbf{\textcolor{red}{0.020}} & \textbf{\textcolor{red}{0.963}} \\

V    & No. IV w/o Transformer & 36.03 & 38.73 & 24.13
    & 0.910 & 0.855 & 0.871 & 0.016 & 0.957 
    & 0.919 & 0.893 & 0.902 & 0.022 & 0.959 \\

VI   & No. IV w/o HDBSCAN     & 40.01 & 39.84 & 12.30
    & 0.912 & 0.857 & 0.871 & 0.017 & 0.959 
    & 0.920 & 0.892 & 0.886 & 0.022 & 0.960 \\

VII  & No. IV w/o $\mathcal{L}_c$ & 40.01 & 39.84 & 22.68
    & 0.909 & 0.878 & 0.886 & 0.018 & 0.960 
    & 0.920 & 0.895 & 0.886 & 0.021 & 0.962 \\
\bottomrule
\end{tabular}

}
    \label{tab:modules_comparison}
\end{table}

\begin{figure}[!t]
    \centering
    \includegraphics[width=\linewidth]{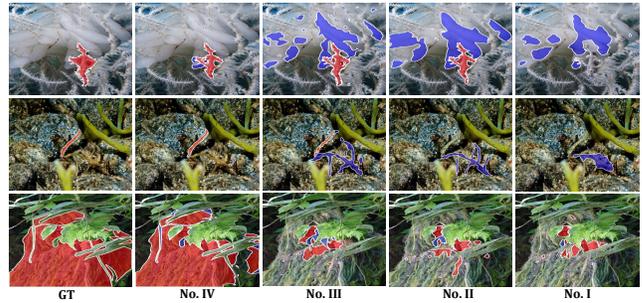}
    \caption{Visual comparison of the variants No. I-IV shown in Tab.~\ref{tab:modules_comparison}.
    The \textcolor{red}{red} and \textcolor{blue}{blue} regions represent correct and error predictions, respectively.}
    \label{fig:visual_mou}
\end{figure}

\begin{table}[!t]
    \centering
  \caption{Comparison of different parameter-efficient finetuning methods based on No. II in Tab.~\ref{tab:modules_comparison}. All GPU memory values are measured with a batch size of 10.}
    \resizebox{\linewidth}{!}{\begin{tabular}{l|c|ccccc|ccccc}
\toprule
\multirow{2}{*}{{Methods}} & \multirow{2}{*}{{GPU Mem (GB)}} 
& \multicolumn{5}{c|}{{COD10K (2,026 images)}} 
& \multicolumn{5}{c}{{NC4K (4,121 images)}} \\
& & $S_\alpha \uparrow$ & $F^w_\beta \uparrow$ & $F_{\beta} \uparrow$ & $M \downarrow$ & $E_m \uparrow$
  & $S_\alpha \uparrow$ & $F^w_\beta \uparrow$ & $F_{\beta} \uparrow$ & $M \downarrow$ & $E_m \uparrow$ \\
\midrule
Full-Tuning                    & 18G  & 0.872 & 0.822 & 0.833 & 0.026 & 0.925 
& 0.889 & 0.836 & 0.849 & 0.031 & 0.935 
 \\
Frozen-Backbone               & 1G   & 0.780 & 0.693 & 0.702 & 0.054 & 0.810 & 0.722 & 0.660 & 0.675 & 0.065 & 0.761 \\
Series-Adapter~\cite{houlsby2019adapter}     & 4G   & 0.840 & 0.756 & 0.771 & 0.026 & 0.874 & 0.835 & 0.780 & 0.792 & 0.036 & 0.875 \\
Parallel-Adapter~\cite{pfeiffer2021adapterfusion} & 4G   & 0.847 & 0.763 & 0.776 & 0.025 & 0.880 & 0.855 & 0.801 & 0.810 & 0.034 & 0.894 \\
ViT-Adapter~\cite{chen2022vitadapter}        & 4G   & 0.863 & 0.777 & 0.790 & 0.025 & 0.895 & 0.878 & 0.825 & 0.835 & 0.033 & 0.917 \\
LoRA~\cite{hu2021lora}                       & 2G   & 0.831 & 0.746 & 0.756 & 0.028 & 0.865 & 0.812 & 0.755 & 0.770 & 0.037 & 0.854 \\
DMA (Ours)                                   & 3G   & \textbf{\textcolor{red}{0.881}} & \textbf{\textcolor{red}{0.833}} & \textbf{\textcolor{red}{0.844}} & \textbf{\textcolor{red}{0.023}} & \textbf{\textcolor{red}{0.930}} 
                                            & \textbf{\textcolor{red}{0.898}} & \textbf{\textcolor{red}{0.848}} & \textbf{\textcolor{red}{0.860}} & \textbf{\textcolor{red}{0.030}} & \textbf{\textcolor{red}{0.941}} \\
\bottomrule  
\end{tabular}}
    \label{tab:adapter_table}
\end{table}

\noindent\textbf{Effectiveness of DMA.}
As shown in Tab.~\ref{tab:modules_comparison}, the effectiveness of the DMA can be validated by comparing variants No. I and II. Efficient parameter fine-tuning introduces only a small number of trainable parameters while significantly improving model performance. 
Transitioning from No. I to No. II leads to notable improvements across all evaluation metrics: the $M$ value increases by an average of 52\%, the $F^{\omega}_{\beta}$ value by 22.1\%, the $F^{}_{\beta}$ value by 24.3\%, and the $E_m$  value by 15.7\%. 
These results demonstrate that the DMA effectively enhances the performance of DINOv2-based models in the COD task.
Furthermore, as illustrated in Fig~\ref{fig:visual_mou}, the visual comparison between No. I and II clearly shows that our DMA enables the model to acquire multi-scale information, thereby locating objects more accurately in complex environments. 
And we compare our method with four representative fine-tuning approaches, including Series-Adapter~\cite{houlsby2019adapter}, Parallel-Adapter~\cite{pfeiffer2021adapterfusion}, LoRA~\cite{hu2021lora}, and ViT-Adapter~\cite{chen2022vitadapter}. 
All methods adopt the same pretrained backbone and initialization weights to ensure a fair comparison.
As shown in Tab.~\ref{tab:adapter_table}, \methodname{} consistently outperforms all baselines across both COD10K and NC4K benchmarks, achieving the best results in all five evaluation metrics.
This demonstrates the superiority of our densely-connected multi-scale design in enhancing camouflaged object perception with minimal computational overhead. 
We also supplement more ablation studies of the DMA in the \supp{} (Sec. B.2).

\noindent\textbf{Effectiveness of IPR.}
The variant No. III extends the learning stage model (No. II) by incorporating the IPR. 
This enhancement leads to average performance gains of 13.5\% and 8.75\% on COD10K and NC4K, respectively, when transitioning from No. II to No. III. 
These improvements indicate that No. III enhances the semantic perception capabilities for camouflaged objects in most scenes.
Despite these advancements, no DMM is incorporated to complement the IPR, which functions solely as a feature fusion module.
Consequently, the inference capabilities of \methodname{} are not fully developed.
This limitation is evident in the visual results of No. III shown in Fig.~\ref{fig:visual_mou}, where the model fails to detect camouflaged objects.
The observation indicates that No. III lacks sufficient generalization ability in challenging camouflage scenes.

\noindent\textbf{Effectiveness of DMM.}
We validate the effectiveness of our memory mechanism (DMM) through both visualization (Fig.~\ref{fig:visual_mou}) and quantitative comparisons (Tab.~\ref{tab:modules_comparison}). 
The variant No. IV demonstrates superior localization accuracy and noise suppression compared to the baseline variant. 
This improvement is achieved through the IPR's reconstruction of sample-adaptive inference patterns that strategically utilize knowledge from the memory mechanism. 
The collaboration of the IPR and the DMM significantly improves the model's capability to handle both unseen and rarely encountered camouflage scenes.  
Moreover, the results of No. V and No. VI in Tab.~\ref{tab:modules_comparison} further demonstrate the superiority of our memory mechanism design. 
Specifically, in No. VI, we remove HDBSCAN~\cite{mcinnes2017hdbscan} and the prototype computation process, opting instead to memory mechanism by computing the similarity between each training sample and the current sample. 
Compared to No. IV, No. VI suffers a 50\% decrease in inference speed, and the memory storage increases from 150\,KB to 8\,MB (90\% increase), accompanied by an average performance drop of 6.25\%. 
In more detail, we visualize the features before and after the dynamic memory mechanism through t-SNE, as shown in Fig.~\ref{fig:cluster}. 
We also provide more analysis on the memory in the \supp{} (Sec. B.3). 

\begin{table}[!t]
    \centering
    \caption{Ablation comparison of clustering algorithm.}
    \resizebox{0.8\linewidth}{!}{\begin{tabular}{l|c|ccccc|ccccc}
\toprule
\multirow{2}{*}{{Settings}} & \multirow{2}{*}{FPS} &
\multicolumn{5}{c|}{{COD10K (2,026 images)}} &
\multicolumn{5}{c}{{NC4K (4,121 images)}} \\
& &
$S_{\alpha}\uparrow$  & $F^{\omega}_{\beta} \uparrow$ & $F_{\beta} \uparrow$ & $\mathcal{M} \downarrow$ & $E_m \uparrow$ &
$S_{\alpha}\uparrow$  & $F^{\omega}_{\beta} \uparrow$ & $F_{\beta} \uparrow$ & $\mathcal{M} \downarrow$ & $E_m \uparrow$ \\
\midrule
Direct Match & 12.30 & 0.906 & 0.857 & 0.868 & 0.017 & 0.952 & 0.911 & 0.882 & 0.893 & 0.024 & 0.960 \\
K-Means      & 30.63 & 0.902 & 0.847 & 0.862 & 0.016 & 0.949 & 0.914 & 0.874 & 0.883 & 0.023 & 0.953 \\
DBSCAN       & 24.12 & 0.910 & 0.862 & 0.871 & 0.015 & 0.958 & 0.917 & 0.893 & 0.902 & 0.021 & 0.957 \\
HDBSCAN      & {22.68} 
             & \textbf{\textcolor{red}{0.915}} & \textbf{\textcolor{red}{0.868}} & \textbf{\textcolor{red}{0.886}} & \textbf{\textcolor{red}{0.014}} & \textbf{\textcolor{red}{0.966}} 
             & \textbf{\textcolor{red}{0.923}} & \textbf{\textcolor{red}{0.900}} & \textbf{\textcolor{red}{0.910}} & \textbf{\textcolor{red}{0.020}} & \textbf{\textcolor{red}{0.963}} \\
\bottomrule
\end{tabular}
}
    \label{tab:cluster_algorithm}
\end{table}


\noindent\textbf{Selection of Clustering Algorithm.}
To evaluate the role of clustering algorithms in the DMM, we compare three popular algorithms in Tab.~\ref{tab:cluster_algorithm}:
K-Means~\cite{Kmeans},
DBSCAN~\cite{DBSCAN},
and HDBSCAN~\cite{mcinnes2017hdbscan},
along with a simple ``Direct Match'' baseline that computes similarity without clustering.
Since K-Means requires a pre-defined number of clusters, it becomes less suitable for COD, where object classes are highly diverse and not explicitly annotated. To address this, we approximate the number of camouflaged object classes in the training set and set the number of clusters to 103 accordingly.
Experiments on COD benchmarks reveal that HDBSCAN delivers superior performance and more stable clustering, while Direct Match and K-Means lags due to its lack of adaptive refinement. 
Compared to Direct Match, centralized matching (\eg HDBSCAN, DBSCAN, K-Means) significantly reduces inference cost by compressing the recall space into a small set of representative prototypes. 
Consequently, we select HDBSCAN as the clustering module for generating compact and discriminative memory representations. 

\begin{table}[!t] 
    \centering
    \caption{Comparison on  challenging camouflage scenes.}
    \resizebox{1\linewidth}{!}{


\begin{tabular}{l|ccccc|ccccc|ccccc}
\toprule
\multirow{2}{*}{{Methods}} & 
\multicolumn{5}{c|}{{Seen Scenes (6,109 images)}} & 
\multicolumn{5}{c|}{{Unseen Scenes (364 images)}} &
\multicolumn{5}{c}{{Rare Scenes (91 images)}} \\
& $S_\alpha \uparrow$ & $F^{\omega}_{\beta} \uparrow$ & $F_{\beta} \uparrow$ & $\mathcal{M} \downarrow$ & $E_m\uparrow$ 
& $S_\alpha \uparrow$ & $F^{\omega}_{\beta} \uparrow$ & $F_{\beta} \uparrow$ & $\mathcal{M} \downarrow$ & $E^m_\phi \uparrow$
& $S_\alpha \uparrow$ & $F^{\omega}_{\beta} \uparrow$ & $F_{\beta} \uparrow$ & $\mathcal{M} \downarrow$ &$E_m\uparrow$ \\
\midrule

SINet~\cite{SINet}     & 0.797 & 0.693 & 0.703 & 0.054 & 0.868 & 0.786 & 0.680 & 0.690 & 0.051 & 0.858 & 0.477 & 0.095 & 0.096 & 0.107 & 0.588 \\
PFNet~\cite{PFNet}     & 0.820 & 0.719 & 0.730 & 0.050 & 0.884 & 0.801 & 0.694 & 0.705 & 0.051 & 0.868 & 0.485 & 0.128 & 0.130 & 0.121 & 0.586 \\
BGNet~\cite{BGNet}     & 0.845 & 0.768 & 0.780 & 0.042 & 0.905 & 0.827 & 0.741 & 0.752 & 0.043 & 0.873 & 0.512 & 0.164 & 0.167 & 0.111 & 0.581 \\
ZoomNet~\cite{zoomnet2022}   & 0.849 & 0.768 & 0.780 & 0.039 & 0.894 & 0.826 & 0.733 & 0.744 & 0.043 & 0.875 & 0.522 & 0.176 & 0.179 & 0.095 & 0.608 \\
FEDER~\cite{FEDER2023}       & 0.840 & 0.767 & 0.779 & 0.041 & 0.905 & 0.813 & 0.725 & 0.736 & 0.045 & 0.873 & 0.522 & 0.193 & 0.196 & 0.098 & 0.630 \\
HitNet~\cite{HitNet}         & 0.875 & 0.827 & 0.840 & 0.033 & 0.929 & 0.856 & 0.797 & 0.809 & 0.035 & 0.916 & 0.575 & 0.284 & 0.288 & 0.103 & 0.678 \\
RISNet~\cite{RISNet}         & 0.880 & 0.825 & 0.838 & 0.033 & 0.928 & 0.860 & 0.794 & 0.806 & 0.038 & 0.910 & 0.571 & 0.273 & 0.277 & 0.112 & 0.620 \\
FSEL~\cite{FSEL}            & 0.889 & 0.838 & 0.851 & 0.027 & 0.943 & 0.867 & 0.807 & 0.819 & 0.032 & 0.921 & 0.593 & 0.314 & 0.319 & 0.090 & 0.665 \\
CGNet~\cite{zhang2024cgcod}  & 0.900 & 0.854 & 0.867 & 0.024 & 0.954 & 0.882 & 0.819 & 0.831 & 0.029 & 0.926 & 0.630 & 0.356 & 0.361 & 0.073 & 0.715 \\
\methodname{} 
& \textbf{\textcolor{red}{0.918}} & \textbf{\textcolor{red}{0.888}} & \textbf{\textcolor{red}{0.901}} & \textbf{\textcolor{red}{0.019}} & \textbf{\textcolor{red}{0.964}}  
& \textbf{\textcolor{red}{0.901}} & \textbf{\textcolor{red}{0.858}} & \textbf{\textcolor{red}{0.871}} & \textbf{\textcolor{red}{0.022}} & \textbf{\textcolor{red}{0.942}}  
& \textbf{\textcolor{red}{0.733}} & \textbf{\textcolor{red}{0.540}} & \textbf{\textcolor{red}{0.548}} & \textbf{\textcolor{red}{0.042}} & \textbf{\textcolor{red}{0.815}} \\

        \bottomrule
    \end{tabular}
}
    \label{tab:seen}
\end{table}

\begin{figure}[!t]
    \centering
    \includegraphics[width=0.8\linewidth]{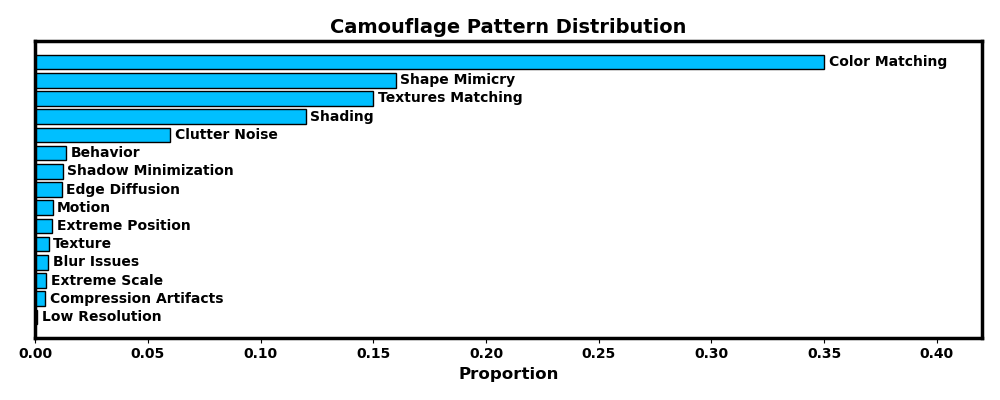}
    \caption{Camouflage pattern distribution of training data.}
    \label{fig:c_pattern}
\end{figure}

\begin{figure}[!t]
\centering  
\includegraphics[width=0.8\linewidth]{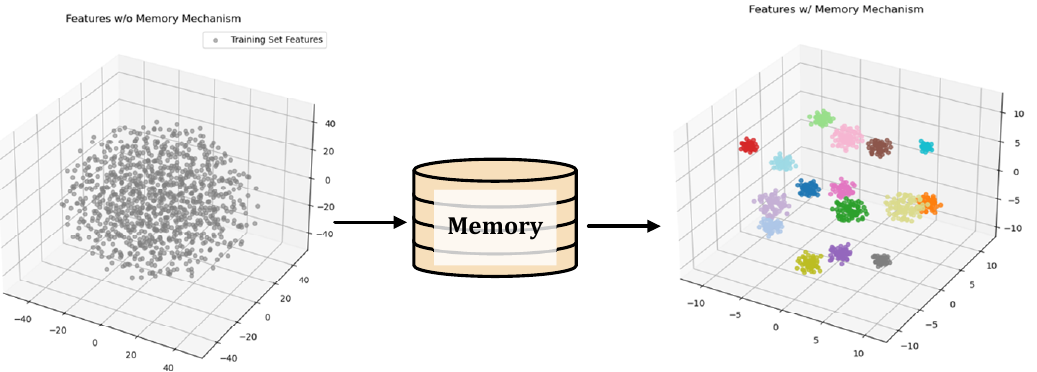}
\caption{Details of the t-SNE method.}

\label{fig:cluster}
\end{figure}

\noindent\textbf{Generalization Ability.}
To evaluate the generalization ability of COD methods, we categorize all data from image COD testing sets including CAMO, COD10K, NC4K, and CHAMELEON, and several representative samples from the video COD dataset MoCA-Mask~\cite{Moca} into seen, unseen, and rare scenes. 
As shown in Fig.~\ref{fig:c_pattern}, samples with camouflage patterns whose training samples account for less than 5\% of the total training set are identified as {rare scenes}. 
Unseen scenes contain samples that do not include any object classes present in the training set. 
This facilitates a more comprehensive evaluation for model generalization and robustness in challenging real-world scenes. 
As shown in Tab.~\ref{tab:seen}, \methodname{} consistently outperforms all comparison methods on these datasets. 
Compared with the latest SOTA method, CGNet~\cite{zhang2024cgcod}, our network achieves performance gains of 7\%, 10\% and 31\% on these three sets, with an even more significant improvement in the rare scene.
This advantage primarily stems from the synergy between our IPR and DMM, allowing the model to incorporate historical information and  reconstruct an adaptive inference pattern for each sample. 
Consequently, the proposed \methodname{} exhibits stronger generalization ability in challenging camouflage scenes.

\section{Conclusion}
\label{sec:conclusion}
We propose \textbf{\methodname}, a novel two-stage COD framework comprising a learning stage and a recall stage. 
By the collaboration of these two stages, the system establishes a context-aware inference mechanism capable of maintaining robust performance across diverse camouflage scenes. 
\methodname{} alleviates the existing limitations of static inference patterns and reliance on instantaneous visual features through the synergistic integration of sample representations and recalled historical knowledge. 
Extensive experiments conducted on the widely-used conventional COD datasets and carefully-constructed
challenging benchmarks demonstrate that our \methodname{} achieves superior performance compared to other state-of-the-art methods. 

\bibliographystyle{ACM-Reference-Format}
\bibliography{sample-base}








\end{document}